\documentclass[lettersize,journal]{IEEEtran}
\usepackage{amsmath,amsfonts}
\usepackage{amssymb}
\usepackage{amsthm}
\usepackage{dsfont}
\usepackage{algorithmic}
\usepackage{algorithm}
\usepackage{array}
\usepackage[caption=false,font=normalsize,labelfont=sf,textfont=sf]{subfig}
\usepackage{stfloats}
\usepackage{url}
\usepackage{verbatim}
\usepackage{graphicx}
\usepackage{cite}

\usepackage[caption=false]{subfig}
\usepackage{color}
\usepackage{dsfont}
\usepackage{booktabs}
\usepackage{bm}
\usepackage{multirow}
\usepackage{multicol}
\usepackage[abs]{overpic}
\usepackage{textcomp}
\usepackage{pifont}
\usepackage[table]{xcolor}
\def\eg{\emph{e.g.,}}

\def\ie{\emph{i.e.,}}

\begin{document}

\title{Harnessing Vision-Language Pretrained Models with Temporal-Aware Adaptation for Referring Video Object Segmentation}

\author{Zikun~Zhou,
        Wentao~Xiong,
        Li~Zhou,
        Xin~Li,\\
        Zhenyu~He,~\IEEEmembership{Senior Member, IEEE}
        and Yaowei~Wang,~\IEEEmembership{Member, IEEE,}
\thanks{
    Zikun Zhou and Zhenyu He are with the School of Computer Science and Technology, Harbin Institute of Technology, Shenzhen, China, and also with the Pengcheng Laboratory, Shenzhen, China (e-mail: zhouzikunhit@gmail.com; zhenyuhe@hit.edu.cn).
    
    Wentao Xiong and Li Zhou are with the School of Computer Science and Technology, Harbin Institute of Technology, Shenzhen, China (e-mail: 21s151086@stu.hit.edu.cn; lizhou.hit@gmail.com).
    
    Xin Li and Yaowei Wang are with the Pengcheng Laboratory, Shenzhen, China. (e-mail: xinlihitsz@gmail.com; wangyw@pcl.ac.cn)
    Zikun Zhou and Wentao Xiong contribute equally to this work.
    }
}

\markboth{Journal of \LaTeX\ Class Files,~Vol.~14, No.~8, August~2021}%
{Shell \MakeLowercase{\textit{et al.}}: A Sample Article Using IEEEtran.cls for IEEE Journals}


\maketitle

\begin{abstract}
The crux of Referring Video Object Segmentation (RVOS) lies in modeling dense text-video relations to associate abstract linguistic concepts with dynamic visual contents at pixel-level. Current RVOS methods typically use vision and language models pretrained independently as backbones. As images and texts are mapped to uncoupled feature spaces, they face the arduous task of learning Vision-Language~(VL) relation modeling from scratch. Witnessing the success of Vision-Language Pretrained (VLP) models, we propose to learn relation modeling for RVOS based on their aligned VL feature space. Nevertheless, transferring VLP models to RVOS is a deceptively challenging task due to the substantial gap between the pretraining task (static image/region-level prediction) and the RVOS task (dynamic pixel-level prediction). To address this transfer challenge, we introduce a framework named VLP-RVOS which harnesses VLP models for RVOS through temporal-aware adaptation. We first propose a temporal-aware prompt-tuning method, which not only adapts pretrained representations for pixel-level prediction but also empowers the vision encoder to model temporal contexts. We further customize a cube-frame attention mechanism for robust spatial-temporal reasoning. Besides, we propose to perform multi-stage VL relation modeling while and after feature extraction for comprehensive VL understanding. Extensive experiments demonstrate that our method performs favorably against state-of-the-art algorithms and exhibits strong generalization abilities. 
\end{abstract}


\section{Introduction}

\IEEEPARstart{R}{eferring} Video Object Segmentation (RVOS) aims to segment the target object in a video according to the referring expression. It has a wide range of applications, including language-based robot controlling~\cite{CMPC,wang2019reinforced}, augmented reality~\cite{r2vos}, and video editing~\cite{LBDT, M3L}. 
As language descriptions inherently exhibit flexibility and diversity, RVOS necessitates comprehensive Vision-Language (VL) understanding abilities to accurately discover and segment the target object.

\begin{figure}[t]
\centering
\includegraphics[width=1\columnwidth]{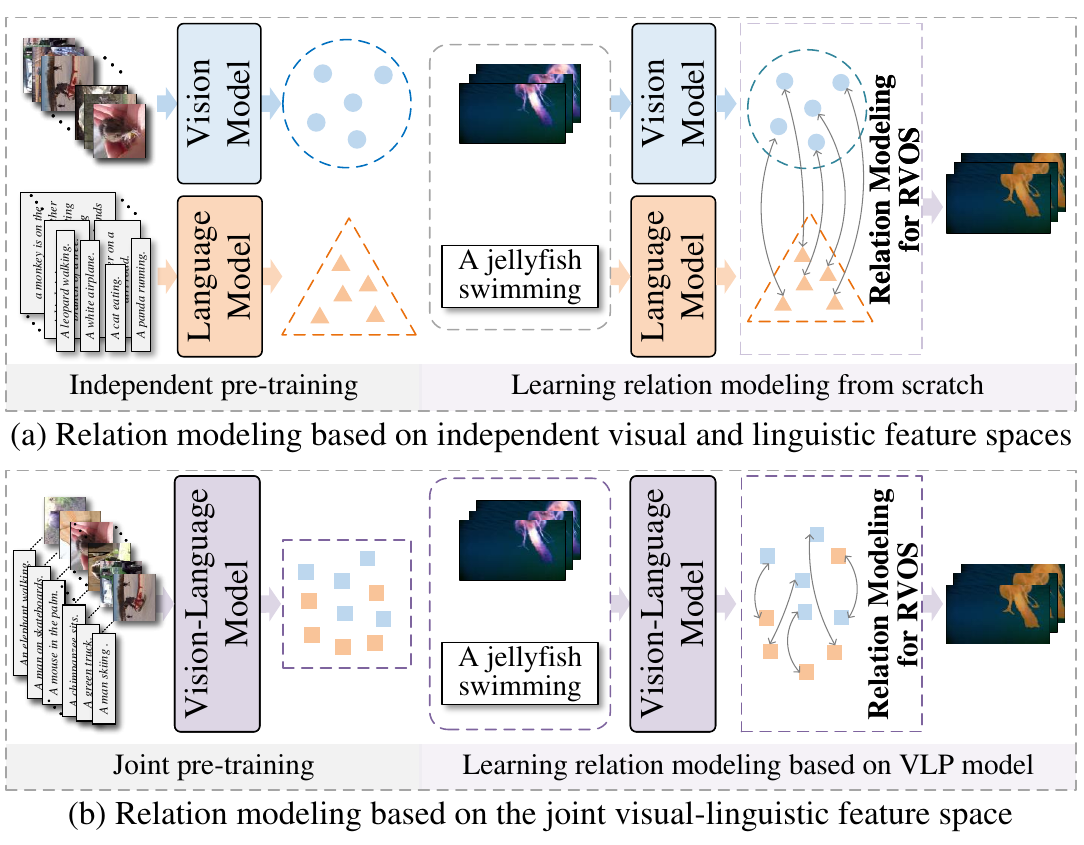}
    \caption{\textbf{Two paradigms of learning dense text-video relation modeling for RVOS.} Compared with learning from scratch, learning such a relation modeling ability based on the aligned VL feature space is more accessible and derives better performance.}
\label{Fig:motivation}
\end{figure}

The crux of RVOS lies in modeling dense text-video relations to associate the diverse yet abstract linguistic concepts with dynamic visual contents at pixel-level. Massive endeavors~\cite{MTTR, ReferFormer,r2vos} have been made for this purpose in the RVOS community. Existing RVOS algorithms~\cite{MTTR, ReferFormer,r2vos} typically build relation modeling components on independently pretrained vision and language backbones, including ResNet~\cite{ResNet}, Video-Swin~\cite{Video-Swin}, and RoBERTa~\cite{RoBERTa}. Such a paradigm for learning relation modeling can be summarized as Figure~\ref{Fig:motivation}~(a). 
As the backbones map input images and texts into decoupled feature spaces, these algorithms face the challenge of learning VL relation modeling for RVOS from scratch. Although incorporating sophisticated relation modeling mechanisms, they struggle to understand complicated descriptions and videos.

\begin{figure}[t]
\centering
\includegraphics[width=1.0\columnwidth]{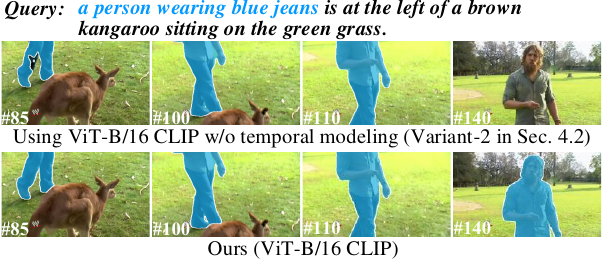}
\vspace{-5mm}
    \caption{\textbf{Comparison between using ViT-B/16 CLIP w/o and w/ temporal modeling.} When the blue jeans disappear from view in the $140^{th}$ frame, our method can still understand that the person is the referred target according to the temporal clue, while the variant without temporal modeling cannot.}
\label{Fig:motivation_compare}
\end{figure}

Recently, Vision-Language pretrained (VLP) models, such as CLIP~\cite{CLIP} and VLMo~\cite{VLMo}, which map images and texts into aligned feature space, have drawn much attention. They have been pivotal in advancing various tasks, such as zero-shot classification~\cite{CLIP} and referring image segmentation~\cite{Cris, ETRIS}. Nevertheless, the application of VLP models in RVOS remains unexplored. In light of this, we seek to unleash the power of VLP models for RVOS, allowing us to learn robust relation modeling for RVOS based on the aligned VL features instead of learning from scratch, as shown in Figure~\ref{Fig:motivation}~(b). Compared with the transfer to image segmentation~\cite{Cris, Denseclip}, the transfer to RVOS poses a more formidable challenge due to the significant gap between the pretraining task (static image/region-level prediction) and the RVOS task (dynamic pixel-level prediction). Particularly, the transfer to RVOS demands not only adapting the image/region-level representation for pixel-level prediction, but also empowering the VLP models with temporal modeling ability. As shown in Figure~\ref{Fig:motivation_compare}, a model using CLIP~\cite{CLIP} without temporal modeling loses the target person when the blue jeans disappear in the $140^{th}$ frame.

In this work, we present a framework called VLP-RVOS which harnesses VLP models for RVOS through temporal-aware adaptation. Specifically, it transfers the knowledge embedded in VLP models to learn robust spatial-temporal and vision-language relation modeling for RVOS. The primary challenge is to learn the above task-specific knowledge from limited video data without forgetting the pretrained knowledge of VL association. To address the issue, we resort to parameter-efficient prompt-tuning, which keeps the VLP model frozen to retain pretrained knowledge and incorporates additional prompts to learn task-specific knowledge. Particularly, we propose a temporal-aware VL prompt-tuning method, which not only adapts the pretrained VL features for pixel-level prediction but also empowers the vision encoder to capture temporal contexts. We also introduce a cube-frame attention mechanism to further facilitate spatial-temporal reasoning for RVOS. Additionally, to ensure comprehensive VL understanding, our framework integrates multi-stage VL relation modeling, including 1) leveraging the linguistic reference to guide visual feature extraction, 2) fusing the deep VL features after feature extraction, and 3) incorporating VL relation modeling during spatial-temporal reasoning.

\begin{figure}[t]
\centering
\includegraphics[width=1.0\columnwidth]{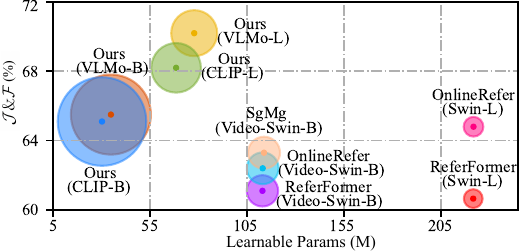}
    \vspace{-5mm}
    \caption{\textbf{Comparison with state-of-the-art algorithms on Ref-DAVIS17~\cite{URVOS}.} We visualize $\mathcal{J}\&\mathcal{F}$ \emph{w.r.t.} the learnable params of different methods. Note that we freeze the VLP model. The circle size indicates the ratio of $\mathcal{J}\&\mathcal{F}$ to the learnable params.}
\label{Fig:Intro_davis}
\vspace{-2mm}
\end{figure}

Extensive experiments on five benchmarks~\cite{A2D_JHMDB,VOS_LRE,URVOS,MeViS} show that VLP-RVOS performs favorably against state-of-the-art methods. Figure~\ref{Fig:Intro_davis} illustrates the comparison in learnable param and $\mathcal{J}\&\mathcal{F}$ on Ref-DAVIS17~\cite{URVOS}. Experimental results show that our framework effectively unleashes the power of VLP to RVOS. 
Our contributions can be concluded as: 
\begin{itemize}
\item We present the VLP-RVOS framework harnessing VLP models for RVOS through temporal-aware adaptation. To the best of our knowledge, this is the first framework designed to facilitate robust VL relation modeling for the RVOS task using the VLP models.

\item We propose a temporal-aware prompt-tuning method, which not only adapts pretrained VL features for pixel prediction but also enables the vision encoder to capture temporal contexts.

\item We tailor a cube-frame attention mechanism to facilitate spatial-temporal reasoning for RVOS and propose a multi-stage VL relation modeling scheme for comprehensive VL understanding.
\end{itemize}

\section{Related Work}
\noindent\textbf{Referring video object segmentation.}
The main challenge of RVOS lies in modeling the dense text-video relation.
Numerous sophisticated VL relation modeling mechanisms~\cite{VTcapsule, MANet, YOFO, PMINet, r2vos, CITD, ReferFormer, CMSA, PRPE} have been proposed to address the challenge.
For example, VT-Capsule~\cite{VTcapsule} uses capsules to model VL representations and fuses the visual and linguistic capsules with a routing mechanism to segment the target. 
Recently, many RVOS algorithms~\cite{MTTR, LBDT, CSTM, ReferFormer, onlinerefer, r2vos, TCE-RVOS} resort to attention-based methods for VL or spatial-temporal relation modeling. Specifically, LBDT~\cite{LBDT} proposes a language-bridged duplex transfer module to accomplish spatial-temporal interaction.
MTTR~\cite{MTTR} and ReferFormer~\cite{ReferFormer} introduce the DETR~\cite{DETR} architecture to RVOS and use language as queries to attend to the referred target.
HTML~\cite{HTML} and TempCD~\cite{TempCD} improve the temporal modeling ability by hybrid temporal-scale learning and temporal collection and distribution, respectively, which achieve promising RVOS performance.

Nevertheless, these RVOS algorithms construct the relation modeling components on independently pretrained vision and language backbones and learn relation modeling from scratch, which is a tough learning task. 
Unlike these methods, we propose to transfer the powerful VLP model to RVOS, allowing us to learn relation modeling for RVOS from a joint VL feature space instead of learning from scratch.

\vspace{1mm}
\noindent\textbf{Referring image segmentation.}
Referring Image Segmentation (RIS) is closely related to RVOS, whose goal is to segment the target object described by the referring expression in a static image~\cite{qiu2019referring_TMM,cho2023cross_TMM}. Similar to existing RVOS algorithms, numerous RIS approaches~\cite{lin2021structured_TMM,liu2022instance_TMM,MRES} adopt a pipeline of first extracting the visual and linguistic features and then modeling the cross-modality relation based on the unimodal representations for image mask prediction. Most of these algorithms~\cite{liu2022instance_TMM,MRES,Lavt,RRSIS} resort to independently pretrained vision and language backbones and learn VL relation modeling from scratch. A few methods~\cite{Cris,ETRIS,RefMatt} build the RIS framework on top of the vision-language pretrained model, CLIP. Compared with the CLIP-based RIS approaches, transferring VLP models to RVOS is much more challenging due to the larger gap between the pretraining task and the RVOS task.

\vspace{1mm}
\noindent\textbf{Vision-language pretrained models.}
Recently, VLP models~\cite{VLMo, ALIGN, oscar, CLIP, vl-bert}, learning multi-modality representation on large-scale image-text pairs, have attracted much attention. 
They typically adopt a dual-stream~\cite{ALIGN, CLIP, vilbert, LXMERT, Blip} or single-stream~\cite{oscar,vl-bert} encoder structure to extract the visual and linguistic features and align them via cross-modality interaction. 
VLP models have driven the progress of various downstream tasks, such as image-text retrieval~\cite{VLMo, oscar, CLIP}, referring image segmentation~\cite{Cris,ETRIS,RefMatt}, and open-vocabulary detection~\cite{Regionclip}. Nevertheless, the exploitation of VLP models for RVOS has not been explored.  In this paper, we try to overcome the discrepancy between the pretraining and RVOS tasks and take a step towards transferring the powerful VLP models to RVOS.

\begin{figure*}[t]
\centering
    \includegraphics[width=1.0\textwidth]{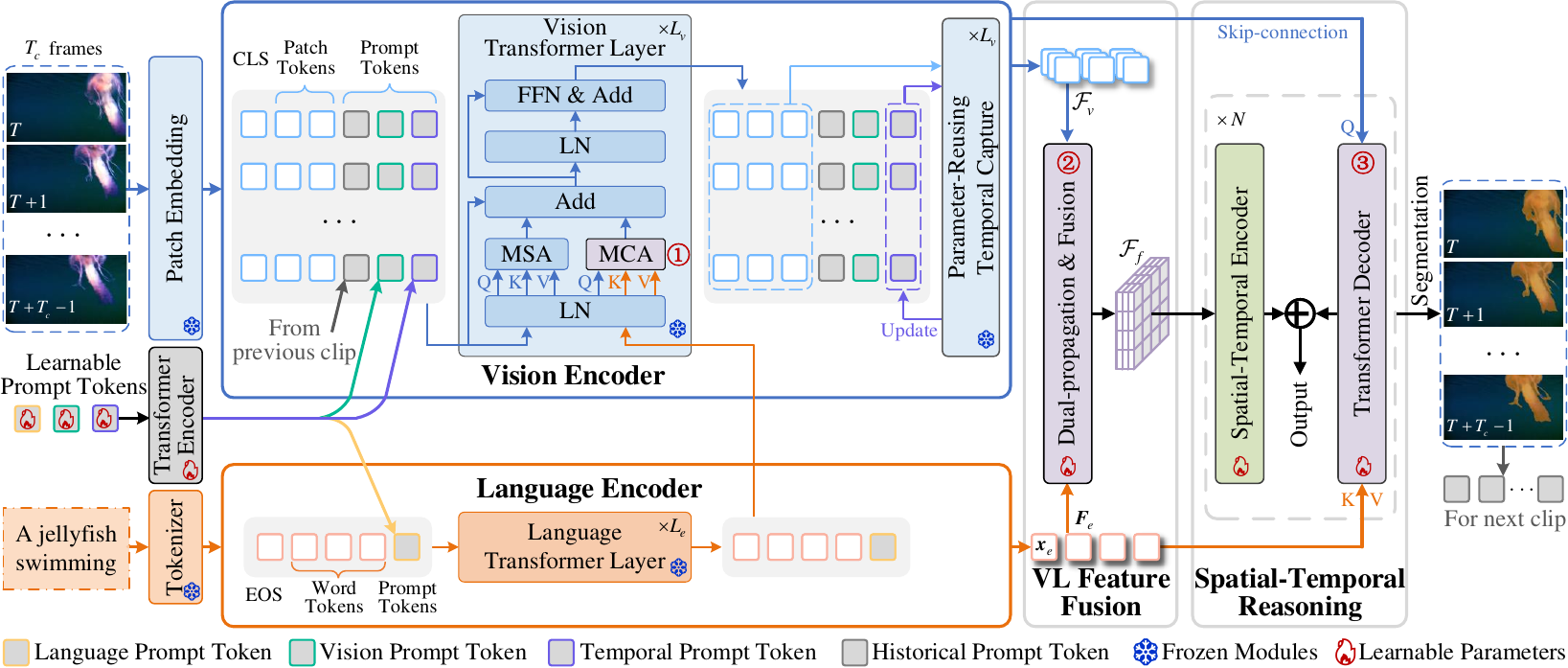}
    \caption{\textbf{Overall architecture of VLP-RVOS,} which processes long videos clip-by-clip. The prompt tokens are first appended to the input VL tokens. Then the vision encoder extracts video features with the guidance of learnable vision/temporal prompts and historical prompts conditioned on the previous clip. The language encoder, tuned by learnable language prompts, extracts linguistic features. VL feature fusion and spatial-temporal reasoning modules associate linguistic concepts with corresponding dynamic visual contents. A segmentation head is used for final target segmentation. \small{\textcircled{\scriptsize{1}}}\small, \small{\textcircled{\scriptsize{2}}}\small~and \small{\textcircled{\scriptsize{3}}}\small~mark the three VL relation modeling stages. MSA/MCA denotes multi-head self/cross-attention. LN is layer normalization. $\oplus$ is element-wise summation.}
    \vspace{-1mm}
\label{Fig:Framework}
\end{figure*}

\vspace{1mm}
\noindent\textbf{Prompt-tuning.} Prompting was proposed in NLP~\cite{Delta-tuning-survey, NLP_prompt1, NLP_pmt} to 
generate task-specific instructions for the language model to obtain desired outputs. Recently, prompt-tuning has been widely explored in vision and multi-modal problems to efficiently adapt the pretrained model to downstream tasks, including image/video recognition~\cite{CoOP, VPT,ni2022expanding}, image segmentation~\cite{ETRIS,kwon2023probabilistic}, video-text retrieval~\cite{Vita-CLIP}, and domain adaptation~\cite{prompt_TMM}. Nevertheless, prompt-tuning has not been explored in the RVOS area, which requires pixel-level video-text understanding and is different from the above-mentioned tasks. In this work, we explore adapting the pretrained VL representation to RVOS via prompt-tuning.

\section{VLP-RVOS}
Figure~\ref{Fig:Framework} illustrates the architecture of VLP-RVOS, which mainly consists of the VLP model, the VL Feature Fusion (VLFF) module, and the Spatial-Temporal Reasoning (STR) module. To learn task-specific knowledge from limited video data without forgetting pretrained knowledge, we opt for parameter-efficient prompt-tuning, instead of fine-tuning the VLP model, which poses the risk of hurting the generalization ability. Specifically, we design a temporal-aware VL prompt-tuning method to enable the vision encoder to capture the temporal context for video understanding. Besides temporal-aware prompt-tuning, we also introduce the STR module to enhance the temporal modeling ability for RVOS further.

For comprehensive VL understanding, VLP-RVOS is designed to conduct three-stage VL relation modeling, marked by the red numbers in Figure~\ref{Fig:Framework}: 1) We introduce additional Multi-head Cross-Attention (MCA) into the vision encoder to leverage the linguistic reference to guide visual feature extraction. 2) We employ the VLFF module to fuse deep VL features for associating high-level visual semantics with abstract linguistic concepts. 3) We perform VL relation modeling between linguistic features and shallow visual features in STR, aiming to introduce low-level semantics to enhance the comprehension of changing visual contents described in the text. Next, we delve into the specifics of VLP-RVOS.

\subsection{Vision-Language (VL) encoders}
\noindent\textbf{Vision encoder.}
Given a clip $\mathcal{V}=\{\bm I^{t}\}_{t=T}^{T+T_{c}-1}$ with $T_{c}$ frames from a long video, where $T$ is the index of its beginning frame, the vision encoder (\eg~ViT-B/16~\cite{ViT} of CLIP) extracts visual features for each frame with the tuning of prompt tokens. We first bracket the patch embeddings of each frame with a $\rm CLS$ token and the prompt tokens, then feed them into the transformer layers for feature extraction. We further process the visual feature with a projection layer to align its dimension $C_{v}$ with that of the linguistic feature $C_{e}$ for dimension consistency. The resulting video feature is denoted by $\mathcal {F}_v\!=\!\{\bm F_v^t\in \mathbb{R}^{(N_v+1) \times C_{e}}\}_{t=T}^{T+T_{c}-1}$, where $\bm F^{t}_{v}$ is the feature of the $t$-th frame and $N_v$ is the number of patch embeddings per frame. Note the output tokens corresponding to prompts are dropped in $\mathcal {F}_v$.

\vspace{1mm}
\noindent\textbf{Language encoder.}
Given a referring expression $\mathcal{E}\!\!=\!\!\{\bm {W}_{n}\}_{n=0}^{N_{w}-1}$ with $N_{w}$ words, we first tokenize each word and bracket the word embedding sequence with an $\rm SOS$ token and an $\rm EOS$ token. Then the language encoder (\eg~the modified Transformer~\cite{Transformer} of CLIP), tuned by learnable language prompts, processes this sequence to extract the linguistic feature $\bm {F}_e\in \mathbb R^{N_e \times C_{e}}$. Herein $N_e$ is the number of linguistic feature tokens. The token in $\bm F_e$ corresponding to $\rm EOS$ is the global representation of $\mathcal{E}$, and we denote it by $\bm x_e$.

\subsection{Temporal-aware VL prompt-tuning}
To preserve pretrained knowledge, we opt for prompt-tuning to adapt the VLP model to RVOS, which keeps the VLP model frozen and learns a small number of prompt tokens. Particularly, we design a temporal-aware VL prompt-tuning method to adapt the VLP model for pixel-level prediction and enable it to capture temporal clues. Next, we elaborate on the prompt-tuning method.

\subsubsection{Temporal-aware vision prompt-tuning}
Prompt-tuning on the vision encoder has two objectives: 1) adapting the visual representation pretrained for image/region-level prediction to pixel prediction; 2) empowering the vision encoder to capture and exploit the temporal context in videos. To this end, we introduce three types of prompt tokens: the vision prompt, the temporal prompt, and the historical prompt.

\vspace{1mm}
\noindent\textbf{Vision prompt.}
The vision prompt tokens $\bm P_{v} \in \mathbb{R}^{M_{v} \times C_{v}}$ are introduced to adapt the pretrained visual representation for pixel prediction. Technically, they are randomly initialized learnable vectors. We adopt a deep prompt-tuning strategy on the vision decoder to provide additional learning capacity for each transformer layer. Specifically, we divide the $M_{v}$ vision prompt tokens into $L_{v}$ groups, each containing $m_{v}=M_{v}/L_{v}$ prompt tokens. These groups are then appended to the patch tokens of each vision transformer layer. Herein $L_{v}$ is the number of vision transformer layers. All frames share the same prompt tokens in each layer.

\begin{figure}[t]
\centering
    \includegraphics[width=1.0\columnwidth]{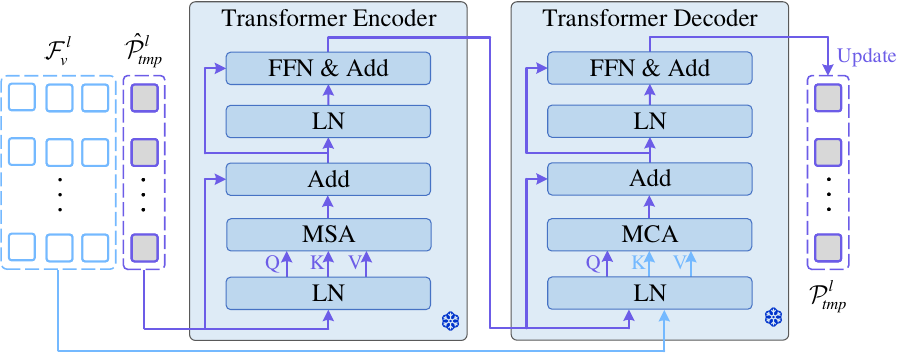}
    \vspace{-2mm}
    \caption{\textbf{Illustration of our Parameter-Reusing Temporal Capture module.} It reuses each transformer layer in the VLP model as the encoder and decoder to capture the temporal clue.}
\label{Fig:temporal_modeling}
\vspace{-3mm}
\end{figure}

\vspace{1mm}
\noindent\textbf{Temporal prompt.}
The temporal prompt tokens $\bm P_{tmp} \in \mathbb{R}^{m_{tmp} \times C_{v}}$ are used as carriers to capture and spread the temporal context in the input video clip. Like $\bm P_{v}$, the temporal prompt tokens $\bm P_{tmp}$ are also randomly initialized learnable vectors. Differently, we adopt a shallow prompt-tuning strategy for the temporal prompt. We repeat $\bm P_{tmp}$ for $T_{c}$ times and append the $t$-th copy $\bm P^{t}_{tmp}$ to the patch embeddings of $\bm I^{t}$ in the first transformer layer. 
In each layer, we use the output embeddings corresponding to the temporal prompt tokens as carriers for temporal modeling. Specifically, we construct a Parameter-Reusing Temporal Capture (PRTC) module on the output of each vision transformer layer to capture the temporal context of the video. In the $l$-th layer, it takes as input the visual features $\mathcal F^{l}_{v}\!=\!\{\bm F_{v}^{t,l}\}_{t=T}^{T+T_{c}-1}$ and temporal embeddings $\hat {\mathcal P}^{l}_{tmp}\!=\!\{\hat {\bm P}_{tmp}^{t,l}\}_{t=T}^{T+T_{c}-1}$ of all frames, and outputs the new temporal embeddings $\mathcal P^{l}_{tmp}\!=\!\{\bm P_{tmp}^{t,l}\}_{t=T}^{T+T_{c}-1}$ modeling the temporal contexts. The temporal contexts embedded in $\mathcal P^{l}_{tmp}$ are further spread to the visual features via the interaction of the $(l\!+\!1)$-th vision transformer layer.

As shown in Figure~\ref{Fig:temporal_modeling}, the PRTC module reuses the frozen visual transformer layer in VLP models as its encoder and decoder. Technically, we directly replace the MSA operation with the MCA operation to convert a transformer encoder into a decoder. PRTC employs the encoder to perform the interaction between the temporal embeddings of all frames and uses the decoder to perform the interaction between the temporal embeddings and visual features of all frames. The temporal contexts are embedded into $\mathcal{P}^{l}_{tmp}$ through the aforementioned cross-frame interactions. Denoting the transformer encoder and decoder by $\Phi^{l}_{Enc}$ and $\Phi^{l}_{Dec}$ in the $l$-th layer, the above operation can be formulated as:
\begin{equation}
\mathcal{P}^{l}_{tmp} = \Phi^{l}_{\rm Dec}(\Phi^{l}_{\rm Enc}(\hat {\mathcal P}^{l}_{tmp}), \mathcal{F}^{l}_{v}).
\end{equation}

\vspace{1mm}
\noindent\textbf{Historical prompt.}
VLP-RVOS processes long videos clip-by-clip. Therefore, we introduce historical prompt tokens, conditioned on the target states in the previous clip, to provide historical prior for the VLP model. Technically, each historical prompt token is calculated by performing masked global pooling and linear projection on the feature of a previous frame with the corresponding mask. We adopt a deep prompt-tuning strategy with the historical prompts by appending them to every vision transformer layer. Particularly, we use different linear projection layers to generate the historical prompt tokens for each visual transformer layer, as different layers have different semantic levels.

With the prompt-tuning method, the processing of the $t$-th frame in the $l$-th vision transformer layer is formulated as:
\begin{align}
    \label{Eq:prompt1}
    & \bm F^{t,l-1}_{p} \!=\! [\bm F_{v}^{t,l-1},\bm P^{l-1}_{h},\bm P^{l-1}_{v},\bm P^{t,l-1}_{tmp}],\\
    \label{Eq:prompt2}
    & \tilde{\bm F}^{t,l-1}_{p} \!=\! \bm F^{t,l-1}_{p} + \phi_{\rm MSA}(\phi_{\rm LN}(\bm F^{t,l-1}_{p})),\\
    \label{Eq:prompt3}
    & [\bm F_{v}^{t,l},\bm P^{l}_{h},\bm P^{l}_{v},\hat {\bm P}^{t,l}_{tmp}] \!=\! \phi_{\rm FFN}(\phi_{\rm LN}(\tilde{\bm F}^{t,l-1}_{p}))\! +\! \tilde{\bm F}^{t,l-1}_{p},
\end{align}
where $\phi_{\rm LN}$, $\phi_{\rm MSA}$, and $\phi_{\rm FFN}$ refer to layer normalization, multi-head self-attention, and feed-forward network in the vision transformer layer.

\subsubsection{Language prompt-tuning}
Language prompt-tuning aims to adapt the pretrained language encoder to understand the referring expression. We append language prompt tokens $\bm P_{e}\!\in\! \mathbb{R}^{m_{e}\times C_{e}}$ to the tokenized word embeddings. These tokens learn the overall distribution of the referring expression data and facilitate the language encoder modeling textual contexts to understand the referring expression comprehensively.

Similar to~\cite{UVLPrompt}, we adopt a transformer encoder to perform the multi-modality prompt interaction before feeding them into the encoders, allowing for the joint learning of multi-modality prompts, as shown in Figure~\ref{Fig:Framework}.

\vspace{-1mm}
\subsection{Multi-stage VL relation modeling}
Herein we present how to perform multi-stage VL relation modeling while and after feature extraction in VLP-RVOS.

\begin{figure}[t]
\centering
    \includegraphics[width=0.9\columnwidth]{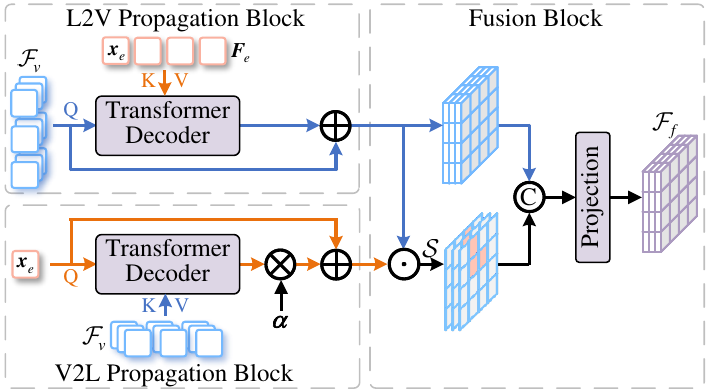}
    \vspace{-2mm}
    \caption{\textbf{Structure of the VL Feature Fusion module,} consisting of a language-to-vision (L2V) propagation block, a vision-to-language (V2L) propagation block, and a fusion block.}
\label{Fig:VLFF}
\vspace{-3mm}
\end{figure}

\vspace{1mm}
\subsubsection{Reference-guided visual encoding during feature extraction}
Unlike many RVOS methods~\cite{ReferFormer,YOFO} performing VL relation modeling only after feature extraction, we propose to inject the linguistic reference information into the visual encoder during feature extraction, serving as the first stage of VL relation modeling. As shown in Figure~\ref{Fig:Framework}, we feed the language feature $\bm F_{e}$ into each vision transformer layer and calculate the cross-attention between $\bm F_{e}$ and the visual embeddings of each layer. To this end, we introduce a Multi-head Cross-Attention (MCA) operation in each vision transformer layer, which reuses the parameter of the existing MSA operation. Owing to the alignment nature between the visual and linguistic features, such a simple parameter-reusing MCA operation can effectively modulate the visual feature with the linguistic concept. The formulation of the attention operation in the $l$-th layer, \ie~Eq.~\eqref{Eq:prompt2}, is modified as follows:
\begin{equation}
\label{Eq:Layer-by-layer}
\begin{split}
    \tilde{\bm F}^{t,l-1}_{p} = &\bm F^{t,l-1}_{p} + \phi_{\rm MSA}(\phi_{\rm LN}(\bm F^{t,l-1}_{p})) + \\
    &\phi_{\rm MCA}(\phi_{\rm LN}(\bm F^{t,l-1}_{p}), \phi_{\rm LN}(\bm F_{e})),
\end{split}
\end{equation}
where $\phi_{\rm MCA}$ denotes multi-head cross-attention.

\vspace{1mm}
\subsubsection{VL feature fusion after feature extraction}
The VL Feature Fusion (VLFF) module, built on the deep VL feature $\mathcal F_{v}$ and $\bm F_{e}$, is used to associate high-level visual semantics with abstract linguistic concepts. As shown in Figure~\ref{Fig:VLFF}, it consists of a vision-to-language (V2L) propagation block, a language-to-vision (L2V) propagation block, and a fusion block. 
The V2L propagation block uses the global linguistic representation $\bm x_{e}$ as the query to calculate MCA with $\mathcal F_{v}$, enhancing the linguistic concepts in $\bm x_{e}$ relevant to the visual content. The L2V propagation block uses $\mathcal F_{v}$ as the query to calculate MCA with the word-level linguistic feature $\bm F_{e}$, enhancing the referred visual contents in $\mathcal F_{v}$. Herein skip connections are introduced for feature stability. Similar to~\cite{EFDL_CLIP}, we rescale the decoder output in the V2L propagation block using a learnable factor $\bm \alpha\! \in \! \mathbb{R}^{C_{e}}$ with small initial values to preserve the alignment between VL features.

With the enhanced VL features, VLFF calculates the pixel-wise cosine similarity between them, obtaining the similarity vectors $\mathcal {S}=\{\bm s^{t}
\in \mathbb{R}^{N_v}\}_{t=T}^{T+T_{c}-1}$. After reshaping the enhanced visual features and similarity vectors into 3D tensors, we concatenate them frame-by-frame for fusion. Finally, we use a projection layer to reduce the dimension of the fusion feature to $C$. We denote the fusion feature by $\mathcal{F}_{f}$.

\begin{figure}[t]
\centering
    \includegraphics[width=1.0\linewidth]{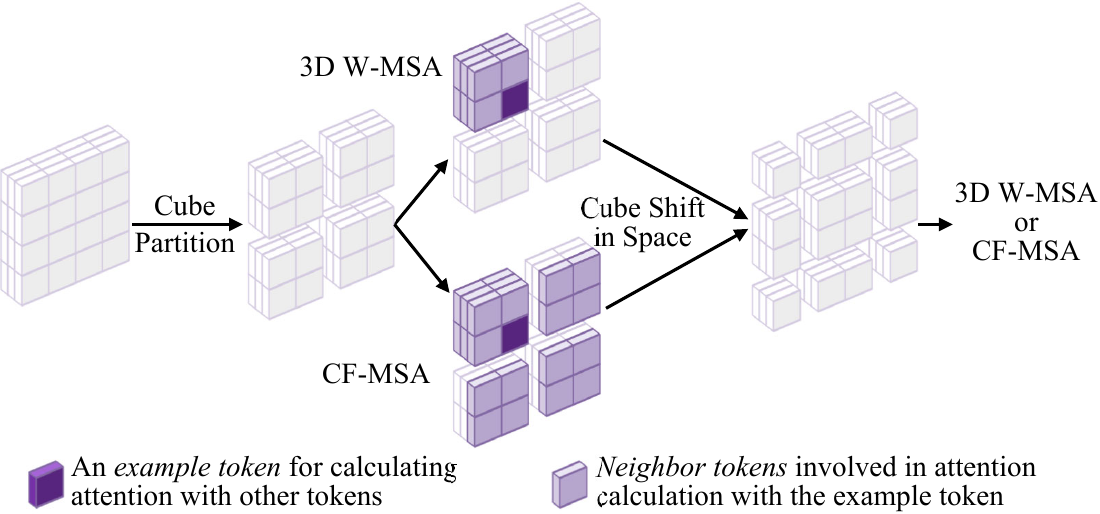}
    \caption{\textbf{Comparison of CF-MSA and 3D W-MSA~\cite{Video-Swin}.} For each token, CF-MSA calculates its attention with its neighbors, including those belonging to the same frame or the same cube. By contrast, 3D W-MSA only calculates the attention within the 3D window. Note that we omit the window partition in the temporal dimension for 3D W-MSA.}
\label{Fig:Attention}
\vspace{-3mm}
\end{figure}

\vspace{1mm}
\subsubsection{VL relation modeling with shallow features}
We further model the VL relation between the shallow visual features and linguistic features in the STR module, which facilitates the STR module associating the changing visual contexts with linguistic concepts. It will be detailed in Section~\ref{Sec: STR}.

\vspace{-1mm}
\subsection{Spatial-temporal reasoning for RVOS}
\label{Sec: STR}
The Spatial-Temporal Reasoning (STR) module aims to capture the dynamic vision contents related to the referring expression, such as objects with varying appearances. As shown in Figure~\ref{Fig:Framework}, it consists of the spatial-temporal encoder and the transformer decoder, which are used to model the spatial-temporal and vision-language relations, respectively, and repeated for $N$ times. 

We devise a Cube-Frame Multi-head Self-Attention (CF-MSA) mechanism for efficient and effective spatial-temporal encoding, as illustrated in Figure~\ref{Fig:Attention}. 
Given the input features $\mathcal F_{f}\in \mathbb R^{T_{c}\times H \times W \times C}$, we first partition it into non-overlap cubes (\ie~3D windows). 
For each token, we calculate its attention with itself and neighbor tokens belonging to the same frame as well as the same cube. Inspired by~\cite{Video-Swin}, we shift the cube in space dimensions and recalculate attention within cubes again for cross-cube modeling. Compared with 3D SW/W-MSA proposed in~\cite{Video-Swin} that models the spatial-temporal relations within 3D windows, CF-MSA further considers the intra-frame global spatial relation, which benefits for perceiving the target location more robustly and accurately. 
Compared with global MSA, CF-MSA omits the relation between two tokens across long spatial and temporal distances, facilitating model learning. Experimental results demonstrate that CF-MSA outperforms global MSA and 3D W-MSA for spatial-temporal encoding, leading to better RVOS performance. 

Assuming the cube size is $T_{c}\!\times\! M_{w}\! \times \!M_{w}$, the computation complexity of the global MSA, 3D W-MSA, and our CF-MSA (w/o cube shift) operations\footnote{Linear Projection and SoftMax are omitted in determining complexity.} on $\mathcal F_{f}\!\in\! \mathbb R^{T_{c}\times H \times W \times C}$ are $\Omega(\text{MSA})=2(T_{c}HW)^{2}C$, $\Omega(\text{3D W-MSA})=2M_{w}^{2}T^{2}_{c}HWC$, and $\Omega(\text{CF-MSA})=2T_{c}HW((T_{c}-1)M_{w}^{2}+HW)C$, respectively. CF-MSA is comparable with 3D W-MSA but surpasses global MSA in efficiency. 

The transformer decoder in STR models the relation between the shallow visual features and the linguistic features, constituting the third stage of VL relation modeling. It introduces additional low-level semantic guidance from the shallow visual features, facilitating STR to understand the variations of the visual contents within a video clip.

\section{Experiments}
\newcommand{\tabincell}[2]{\begin{tabular}{@{}#1@{}}#2\end{tabular}}
\begin{table*}[t]
\begin{center}
\setlength{\tabcolsep}{11pt}
\caption{\textbf{Ablation studies of each component on Ref-Youtube-VOS.} LP, VP, TP, and HP denote the Language Prompt, Vision Prompt, Temporal Prompt, and Historical Prompt, respectively. Stage-1/2/3 denotes the three VL relation modeling stages.}
\vspace{-3mm}
\renewcommand{\arraystretch}{1.}
\resizebox{0.95\linewidth}{!}{
\begin{tabular}{c|ccc|ccc|ccc|c}
\toprule
& \multicolumn{3}{c|}{\emph{Prompt-tuning}} & \multicolumn{3}{c|}{\emph{VL relation modeling}} & \multicolumn{3}{c|}{\emph{Spatial-temporal encoder}} & $\mathcal{J}\&\mathcal{F}$\\
\textcolor{white}{----} & LP+VP & TP & HP & Stage-1 & Stage-2 & Stage-3 & CF-MSA & Global MSA & 3D W-MSA & (\%) \\
\midrule
1) & \textcolor{gray}{\ding{56}} & \textcolor{gray}{\ding{56}} & \textcolor{gray}{\ding{56}} & \textcolor{gray}{\ding{56}} & \textcolor{gray}{\ding{56}} & \ding{52} & \textcolor{gray}{\ding{56}} &\textcolor{gray}{\ding{56}} & \textcolor{gray}{\ding{56}} & 47.9 \\
\midrule
2) & \ding{52} & \textcolor{gray}{\ding{56}} & \textcolor{gray}{\ding{56}} & \textcolor{gray}{\ding{56}} & \textcolor{gray}{\ding{56}} & \ding{52} & \textcolor{gray}{\ding{56}} &\textcolor{gray}{\ding{56}} & \textcolor{gray}{\ding{56}} & 51.8 \\
3) & \ding{52} & \ding{52} & \textcolor{gray}{\ding{56}} & \textcolor{gray}{\ding{56}} & \textcolor{gray}{\ding{56}} & \ding{52} & \textcolor{gray}{\ding{56}} &\textcolor{gray}{\ding{56}} & \textcolor{gray}{\ding{56}} & 54.1 \\
4) & \ding{52} & \ding{52} & \ding{52} & \textcolor{gray}{\ding{56}} & \textcolor{gray}{\ding{56}} & \ding{52} & \textcolor{gray}{\ding{56}} &\textcolor{gray}{\ding{56}} & \textcolor{gray}{\ding{56}} & 54.9 \\
\midrule
5) & \ding{52} & \ding{52} & \ding{52} & \ding{52} & \textcolor{gray}{\ding{56}} & \ding{52} & \textcolor{gray}{\ding{56}} &\textcolor{gray}{\ding{56}} & \textcolor{gray}{\ding{56}} & 56.3 \\
6) & \ding{52} & \ding{52} & \ding{52} & \textcolor{gray}{\ding{56}} & \ding{52} & \ding{52} & \textcolor{gray}{\ding{56}} &\textcolor{gray}{\ding{56}} & \textcolor{gray}{\ding{56}} & 56.0 \\
7) & \ding{52} & \ding{52} & \ding{52} & \ding{52} & \ding{52} & \ding{52} & \textcolor{gray}{\ding{56}} &\textcolor{gray}{\ding{56}} & \textcolor{gray}{\ding{56}} & 57.5 \\
\midrule
8) & \ding{52} & \ding{52} & \ding{52} & \ding{52} & \ding{52} & \ding{52} & \ding{52} &\textcolor{gray}{\ding{56}} & \textcolor{gray}{\ding{56}} & \textbf{59.7} \\
9) & \ding{52} & \ding{52} & \ding{52} & \ding{52} & \ding{52} & \ding{52} & \textcolor{gray}{\ding{56}} & \ding{52} & \textcolor{gray}{\ding{56}} & 58.4 \\
10) & \ding{52} & \ding{52} & \ding{52} & \ding{52} & \ding{52} & \ding{52} & \textcolor{gray}{\ding{56}} & \textcolor{gray}{\ding{56}} & \ding{52} & \underline{58.5} \\
\bottomrule
\end{tabular}}
\label{Tab:Ablation_component}
\end{center}
\vspace{-2mm}
\end{table*}

\subsection{Experimental settings}
\noindent\textbf{Datasets and metrics.} We evaluate VLP-RVOS on Ref-Youtube-VOS~\cite{URVOS}, Ref-DAVIS17~\cite{VOS_LRE}, A2D-Sentences~\cite{A2D_JHMDB}, JHMDB-Sentences~\cite{A2D_JHMDB}, and MeViS~\cite{MeViS}. For Ref-Youtube-VOS, Ref-DAVIS17, and MeViS, region similarity $\mathcal{J}$, contour accuracy $\mathcal{F}$, and their average value $\mathcal{J} \& \mathcal{F}$ are used as metrics, following~\cite{URVOS,MeViS}. For A2D/JHMDB-Sentences, mAP, overall IoU, and mean IoU are used as metrics, following~\cite{A2D_JHMDB}.

\vspace{1mm}
\noindent\textbf{Implementation details.}
We test our algorithm using different VLP models, including ViT-B/16 CLIP, ViT-L/14 CLIP~\cite{CLIP}, VLMo-B, and VLMo-L~\cite{VLMo}. For ViT-B/16 CLIP, we enlarge the input image size from 224 to 352 and interpolate the pretrained positional embeddings. For ViT-L/14 CLIP, VLMo-B, and VLMo-L, we use the original input image sizes, which are 336, 384, and 384, respectively.
$C$ is set to 256 to reduce computation complexity. $N$ is set to 4. 
During training, we freeze the VLP model and optimize the remaining parameters using AdamW~\cite{AdamW} with a weight decay of $5\times 10^{-4}$ and a learning rate of $5\times 10^{-5}$.
Specifically, for Ref-Youtube-VOS, we train the model on its training set alone and report results on its validation set. We also try to pretrain our model on Ref-COCO/+/g~\cite{RefCOCO,RefCOCO+} and fine-tune it on Ref-Youtube-VOS, similar to~\cite{ReferFormer}. 
For Ref-DAVIS17, we directly report the results of the models trained on Ref-Youtube-VOS, providing insights into cross-dataset generalization. For A2D/JHMDB-Sentences, we train our model on the A2D-Sentences training set alone following~\cite{A2D_JHMDB}. For MeViS, we train the model on its training set alone, following~\cite{MeViS}. We use the Dice~\cite{Dice} and Focal~\cite{Focal_loss} losses for end-to-end learning, whose weights are tuned to be 5 and 2, respectively. For image training data~\cite{RefCOCO+,RefCOCO}, we set $T_{c}$ to 1, similar to~\cite{ReferFormer}. For video training data~\cite{A2D_JHMDB,VOS_LRE,URVOS}, we set $T_c$ to 6 and train our model with two consecutive clips sampled from the same videos for each iteration. Thus we can generate historical prompts from the former clip and feed them into the model when performing forward propagation on the latter clip, which allows our VLP-RVOS learning to exploit the historical prior. During inference, $T_{c}$ is set to 6 by default to maintain consistency with the training settings. 
We will release our source codes.

\vspace{-2mm}
\subsection{Ablation studies}
We first conduct ablation studies to analyze our VLP-RVOS framework. We use ViT-B/16 CLIP~\cite{CLIP} as the VLP model and train all the variants on Ref-Youtube-VOS alone for all ablation study experiments.

\begin{table*}[t]
\begin{center}
\setlength{\tabcolsep}{5pt}
\caption{\textbf{Comparisons of different adaptation and temporal modeling methods over our VLP-RVOS framework.} Full fine-tuning means fine-tuning the entire vision encoder. Partial-$m$ means fine-tuning only the last $m$ layers of the vision encoder. PRTC denotes the Parameter-Reusing Temporal Capturing module. $\mathcal{J}\&\mathcal{F}$ is reported.}
\vspace{-3mm}
\renewcommand{\arraystretch}{1.0}
\resizebox{1.0\linewidth}{!}{
\begin{tabular}{l|cccccc|ccc}
\toprule
& \multicolumn{6}{c|}{\emph{Tuning methods}} & \multicolumn{3}{c}{\emph{Temporal modeling methods}}\\
& Frozen & Partial-1 & Partial-3 & \tabincell{c}{Full fine-tuning}  & \tabincell{c}{Adapter-tuning} & \tabincell{c}{\textbf{Prompt-tuning (Ours)}} & TeViT~\cite{TeViT} & IFC~\cite{IFC} & \tabincell{c}{\textbf{PRTC (Ours)}}\\
\midrule 
Ref-Youtube-VOS & 54.3 & 55.1 & \underline{58.5} & 56.8 & 58.0 & \textbf{59.7} & 57.6 & \underline{58.2} & \textbf{59.7} \\
Ref-DAVIS17 & \underline{58.2} & 57.1 & 55.1 & 53.5 & 58.1 & \textbf{60.3} & 58.2 & \underline{58.8} & \textbf{60.3}\\
\bottomrule
\end{tabular}}
\label{Tab:Adaptation_adaptation}
\vspace{-4mm}
\end{center}
\end{table*}

\begin{table*}[h]
\begin{center}
\setlength{\tabcolsep}{13pt}
\caption{\textbf{Experimental results of cross-using VLP models and applying CLIP to other RVOS frameworks.} All models are trained on Ref-Youtube-VOS alone.}
\vspace{-3mm}
\renewcommand{\arraystretch}{1.1}
\resizebox{1.0\linewidth}{!}{
\begin{tabular}{l|c|c|c|cc}
\toprule
\multirow{2}{*}{Algorithms} & Pretrained & Pretrained & Aligned & \multicolumn{2}{c}{$\mathcal{J}\&\mathcal{F}$ (\%)}\\
& Vision Encoder & Language Encoder & VL Space & Ref-Youtube-VOS & Ref-DAVIS17\\
\midrule
ReferFormer~\cite{ReferFormer}+CLIP & CLIP ViT-B/16 & CLIP BERT & \ding{52} & 51.8 & 51.1 \\
SgMg~\cite{SgMg}+CLIP & CLIP ViT-B/16 & CLIP BERT & \ding{52} & 52.7 & 51.9 \\ 
\midrule
Ours (CLIP) & CLIP ViT-B/16 & CLIP BERT & \ding{52} &  \underline{59.7} & \underline{60.3} \\ 
Ours (VLMo) & VLMo-B Vision Encoder & VLMo-B Language Encoder & \ding{52} & \textbf{60.1} & \textbf{61.2}\\ 
Ours (CLIP-VLMo) & CLIP ViT-B/16 & VLMo-B Language Encoder & \textcolor{gray}{\ding{56}} & 55.7 & 52.2  \\ 
Ours (VLMo-CLIP) & VLMo-B Vision Encoder & CLIP BERT & \textcolor{gray}{\ding{56}} & 54.8 & 50.5\\ 
\bottomrule
\end{tabular}}
\label{Tab:source}
\end{center}
\vspace{-4mm}
\end{table*}

\begin{table}[t]
\begin{center}
\setlength{\tabcolsep}{4pt}
\caption{\textbf{Experimental results with varying inference clip lengths of our VLP-RVOS with ViT-B/16 CLIP on Ref-Youtube-VOS.} Please note that videos in Ref-Youtube-VOS are 6 FPS.}
\vspace{-3mm}
\renewcommand{\arraystretch}{1.1}
\resizebox{1.0\linewidth}{!}{
\begin{tabular}{l|ccccccc|c}
\toprule
\tabincell{l}{Clip length\\\textcolor{gray}{Time duration}} & \tabincell{c}{3\\\textcolor{gray}{0.5s}} & \tabincell{c}{6\\\textcolor{gray}{1s}} & \tabincell{c}{12\\\textcolor{gray}{2s}} & \tabincell{c}{18\\\textcolor{gray}{3s}} & \tabincell{c}{24\\\textcolor{gray}{4s}} & \tabincell{c}{30\\\textcolor{gray}{5s}} & \tabincell{c}{36\\\textcolor{gray}{6s}} & \tabincell{c}{Var.}\\ 
\midrule
$\mathcal{J}\&\mathcal{F}$ (\%) & 62.7 & 62.9 & 62.8 & 63.0 & 63.1 & 62.8 & 62.8 & 0.019 \\
FLOPs (G) & 73.27 & 71.91 & 71.25 & 71.04 & 70.94 & 70.89 & 70.87 & --\\
GPU Mem. (MB) & 2,529 & 3,051 & 3,809 & 4,645 & 5,063 & 5,431 & 6,589 & --\\
\bottomrule
\end{tabular}}
\label{Tab:window_size}
\end{center}
\vspace{-3mm}
\end{table}

\subsubsection{Analyses on proposed components}
We analyze the proposed components through 10 variants, as shown in Table~\ref{Tab:Ablation_component}. The experiments begin with a baseline (Variant-1) consisting of a frozen VLP model, $N$ transformer decoder layers originally used for stage-3 VL relation modeling, and a segmentation head.

\vspace{1mm}
\noindent\textbf{Analyses on temporal-aware VL prompt-tuning.}
We gradually introduce different prompts into the baseline
to analyze their effect (Variant-2/3/4). The language and vision prompts enable Variant-2 to adapt pretrained representations for pixel-level prediction, improving $\mathcal{J}\&\mathcal{F}$ by 3.9\%. By introducing the temporal prompts (3,072 learnable parameters) and the PRTC module (no learnable parameters), Variant-3 improves $\mathcal{J}\&\mathcal{F}$ by 2.3\%. It shows that a few learnable parameters effectively enhance the temporal modeling ability. The performance gap between Variant-3 and Variant-4 indicates that historical prompts benefit RVOS in the clip-by-clip inference paradigm.

\vspace{1mm}
\noindent\textbf{Analyses on multi-stage VL relation modeling.}
We introduce the first two stages of VL relation modeling into Variant-4 to analyze our multi-stage VL relation modeling scheme (Variant-5/6/7). The comparisons between Variant-4/5/6/7 manifest that both Stage-1 and Stage-2 contribute to a stronger VL understanding ability, and integrating the three stages further improves RVOS performance.

\vspace{1mm}
\noindent\textbf{Analyses on spatial-temporal attention.}
We construct Variant-8/9/10 modeling the dense spatial-temporal relation with our CF-MSA, global MSA, and 3D SW/W-MSA~\cite{Video-Swin}, respectively. Compared with Variant-7 which only models the temporal context by the vision encoder, all the attention mechanisms bring performance gains, demonstrating the necessity of explicit spatial-temporal relation modeling. Besides, CF-MSA achieves the largest performance gain of 2.2\% among the three attention methods, demonstrating its effectiveness.

\begin{figure}[t]
\centering
\includegraphics[width=1\columnwidth]{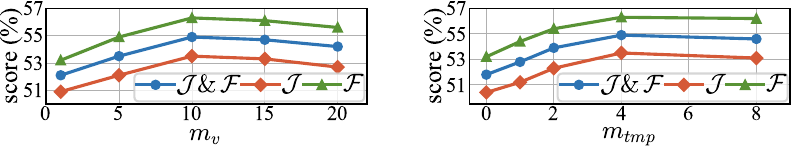}
    \vspace{-7mm}
    \caption{\textbf{Experimental results with varying visual/language prompt token numbers (left) and temporal prompt token numbers (right).}}
\label{Fig:prompt_ablation}
\vspace{-3mm}
\end{figure}

\vspace{1mm}
\subsubsection{Analyses on model tuning methods}
We conduct experiments with several popular model tuning methods on VLP-RVOS to analyze their effect. The involved methods include: 1) Frozen, in which the VLP model is frozen; 2) Partial-$m$, in which the last $m$ layers of the vision encoder are fine-tuned; 3) Full fine-tuning, in which the entire vision encoder is fine-tuned (note that the language encoder is kept frozen following~\cite{ReferFormer,onlinerefer,Denseclip}); 4) Adapter-tuning, in which additional MLP layers are introduced to tune pretrained representations. Herein we use UniAdapter~\cite{UniAdapter}, which has proven to be effective on several cross-modality tasks.

We report the within-dataset (Ref-Youtube-VOS) and cross-dataset (Ref-DAVIS17) evaluation results in Table~\ref{Tab:Adaptation_adaptation}. Although the frozen method obtains the worst performance on Ref-Youtube-VOS, it performs well on Ref-DAVIS17 as it retains the pretrained knowledge. Compared with partial fine-tuning, full fine-tuning cannot obtain better performance and even harm generalization. UniAdapter~\cite{UniAdapter} obtains mediocre performance. We speculate the reason is that it is designed for image-level VL understanding.
Our prompt-tuning performs best on the two benchmarks, demonstrating its effectiveness and generalization.

\begin{table*}[t]
    \centering
    \footnotesize
    \renewcommand\arraystretch{1.02}
    \setlength{\tabcolsep}{8pt}
     \caption{\textbf{Experimental results on Ref-YouTube-VOS and Ref-DAVIS17.} Many RVOS methods use Video-Swin as the visual backbone, while VLP models typically use ViT as the visual encoder. For relatively fair evaluation, we measure the FLOPs per frame and speed of the RVOS models on RTX3090 and split those with similar efficiency to the same group for comparison. Specifically, we compare our models using ViT-B and VLMo-B with those using Video-Swin-T, and compare our models using ViT-L and VLMo-L with those using Video-Swin-B and Swin-L, considering their similar FLOPs and Speed.}
    \resizebox{1.0\textwidth}{!}{
    \begin{tabular}{l|ccc|ccc|ccc}
    \toprule
    \multirow{2}{*}{Algorithms} &\textbf{}
    \multirow{2}{*}{\tabincell{c}{Visual Backbone}} &
    \multirow{2}{*}{\tabincell{c}{FLOPs (G)}} &
    \multirow{2}{*}{\tabincell{c}{Speed (FPS)}} &
    \multicolumn{3}{c|}{Ref-YouTube-VOS} & 
    \multicolumn{3}{c}{Ref-DAVIS17}\\
    &&&&
    \multicolumn{1}{c}{$\mathcal{J}\&\mathcal{F}$ (\%)} &
    \multicolumn{1}{c}{$\mathcal{J}$ (\%)} &
    \multicolumn{1}{c|}{$\mathcal{F}$ (\%)} &
    \multicolumn{1}{c}{$\mathcal{J}\&\mathcal{F}$ (\%)} &
    \multicolumn{1}{c}{$\mathcal{J}$ (\%)} & 
    \multicolumn{1}{c}{$\mathcal{F}$ (\%)} \\
    \midrule
    \multicolumn{10}{l}{\textit{Trained on Ref-Youtube-VOS alone}} \\
    \midrule
    MTTR~\cite{MTTR} & Video-Swin-T & -- & -- & 55.3 & 54.0 & 56.6 & -- & -- & --  \\
    MANet~\cite{MANet} & Video-Swin-T & -- & -- & 55.6 & 54.8 & 56.5 & -- & -- & --   \\
    ReferFormer~\cite{ReferFormer} & Video-Swin-T & 72 & 59 & 56.0 & 54.8 & 57.3 & 55.8 & 51.8 & 59.8   \\   
    SgMg~\cite{SgMg} & Video-Swin-T & 65 & 67 & 58.9 & 57.7 & 60.0 & 56.7 & 53.3 & 60.0\\
    SOC~\cite{SOC} & Video-Swin-T & 43 & 64 & 59.2 & 57.8 & 60.5 & 59.0 & 55.4 & 62.6\\  
    \rowcolor{gray!20}
    \textbf{Ours} (CLIP) & ViT-B/16 & 72 & 77 & \underline{59.7} & \underline{57.9} & \underline{61.5} & \underline{60.3} & \underline{56.7} & \underline{64.0} \\
    \rowcolor{gray!20}
    \textbf{Ours} (VLMo) & VLMo-B & 61 & 55 & \textbf{60.1} & \textbf{58.4} & \textbf{61.8} & \textbf{61.2} & \textbf{57.3} & \textbf{65.1}\\
    \midrule
    \multicolumn{10}{l}{\textit{Pretrained on Ref-COCO/+/g and fine-tuned on Ref-Youtube-VOS}} \\
    \midrule
    R2VOS~\cite{r2vos} & Video-Swin-T & -- & -- & 61.3 & 59.6 & 63.1 & -- & -- & --  \\
    ReferFormer~\cite{ReferFormer} & Video-Swin-T & 72 & 59 & 59.4 & 58.0 & 60.9 & 59.7 & 56.6 & 62.8\\
    SgMg~\cite{SgMg} & Video-Swin-T & 65 & 67 & 62.0 & 60.4 & 63.5 & 61.9 & 59.0 & 64.8 \\
    SOC~\cite{SOC} & Video-Swin-T & 43 & 64 & 62.4 & 61.1 & 63.7 & 63.5 & 60.2 & 66.7 \\
    \rowcolor{gray!20}
    \textbf{Ours} (CLIP) & ViT-B/16 & 72 & 77 & \underline{62.9} & \underline{61.3} & \underline{64.4} & \underline{65.1} & \underline{61.4} & \underline{68.8} \\
    \rowcolor{gray!20}
    \textbf{Ours} (VLMo) & VLMo-B & 61 & 55 & \textbf{63.1} & \textbf{61.5} & \textbf{64.7} & \textbf{65.5} & \textbf{60.7} & \textbf{69.4}   \\
    \midrule
    ReferFormer~\cite{ReferFormer} & Video-Swin-B & 132 & 35 & 62.9 & 61.3 & 64.6 & 61.1 & 58.1 & 64.1\\
    OnlineRefer~\cite{onlinerefer} & Video-Swin-B & 127 & 11 & 62.9 & 61.0 & 64.7 & 62.4 & 59.1 & 65.6    \\
    VLT~\cite{vlt} & Video-Swin-B & -- & -- & 63.8 & 61.9 & 65.6 & 61.6 & 58.9 & 64.3  \\
    HTML~\cite{HTML} & Video-Swin-B & -- & -- & 63.4 & 61.5 & 65.2 & 62.1 & 59.2 & 65.1  \\
    SgMg~\cite{SgMg} & Video-Swin-B & 121 & 41& 65.7 & 63.9 & 67.4 & 63.3 & 60.6 & 66.0 \\
    TempCD~\cite{TempCD} & Video-Swin-B & -- & -- & 65.8 & 63.6 & 68.0 & 64.6 & 61.6 & 67.6  \\
    SOC~\cite{SOC} & Video-Swin-B & 98 & 34 & 66.0 & 64.1 & 67.9 & 64.2 & 61.0 & 67.4 \\
    DsHmp~\cite{DsHmp} & Video-Swin-B & -- & -- & \underline{67.1} & 65.0 & \underline{69.1} & 64.9 & 61.7 & 68.1\\
    ReferFormer~\cite{ReferFormer} & Swin-L & 220 & 37 & 62.4 & 60.8 & 64.0 & 60.5 & 57.6 & 63.4  \\
    HTML~\cite{HTML} & Swin-L & -- & -- & 63.4 & 61.5 & 65.3 & 61.6 & 58.9 & 64.4 \\
    OnlineRefer~\cite{onlinerefer} & Swin-L & 222 & 11 & 63.5 & 61.6 & 65.5 & 64.8 & 61.6 & 67.7 \\
    HTR~\cite{HTR} & Swin-L & -- & -- & \underline{67.1} & \underline{65.3} & 68.9 & 65.6 & 62.3 & 68.8 \\
    \rowcolor{gray!20}
    \textbf{Ours} (CLIP) & ViT-L/14 & 219 & 30 & 66.0 & 63.6 & 68.3 & \underline{68.2} & \underline{64.6} & \underline{71.8} \\
    \rowcolor{gray!20}
    \textbf{Ours} (VLMo) & VLMo-L & 183 & 22 & \textbf{67.6} & \textbf{65.3} & \textbf{69.8} & \textbf{70.2} & \textbf{66.3} & \textbf{74.1}  \\ 
    \bottomrule
    \end{tabular}}
    \label{Tab:YTB}
    \vspace{-2mm}
\end{table*}
\vspace{1mm}
\subsubsection{Analyses on temporal modeling methods}
We conduct experiments with two popular methods also designed for temporal modeling within vision transformers, TeViT~\cite{TeViT} and IFC~\cite{IFC}. TeViT~\cite{TeViT} shifts several learnable tokens across frames for temporal modeling. IFC~\cite{IFC} introduces a trainable transformer encoder for temporal aggregation, introducing 66 M learnable parameters. We evaluate the two methods in VLP-RVOS and report the results in Table~\ref{Tab:Adaptation_adaptation}. The comparisons demonstrate the superiority of our PRTC. Moreover, PRTC occupies a small proportion of the computational load during the visual encoding process. For instance, the total FLOPs per frame for visual encoding with VLMo-L amount to approximately 152.2G, whereas for PRTC, it is only 1.9G.

\vspace{1mm}
\subsubsection{Effect of the prompt token number}
We conduct studies on the number of vision/language prompt tokens ($m_v$ and $m_e$) and the number of temporal prompt tokens ($m_{tmp}$) based on Variant-4. We directly set $m_e\!=\!m_v$ to narrow the hyper-parameter search space. As shown in Figure~\ref{Fig:prompt_ablation}, the performance improves along with $m_v$ and $m_{tmp}$ increasing and saturates at $10$ and $4$, respectively.

\vspace{1mm}
\subsubsection{Effect of the aligned VL space and our transferring framework}
We delve deeper into analyses by breaking the aligned VL space and applying CLIP to existing RVOS frameworks. We break the aligned VL space by cross-using the vision and language encoders of CLIP and VLMo, which results in significant performance drops, as shown in Table~\ref{Tab:source}. We also integrate ViT-B/16 CLIP into ReferFormer and SgMg, where we follow~\cite{ViTDet} to obtain hierarchical features based on ViT. Herein we use the same input image size as our VLP-RVOS, which is 352. Table~\ref{Tab:source} shows that ReferFormer and SgMg using CLIP obtain inferior performance compared with our VLP-RVOS. This implies that both ReferFormer and SgMg fail to fully harness the potential of the pretrained CLIP model for RVOS. Overall, these results highlight that both the aligned VL space and our transferring framework are crucial for VLP-RVOS to achieve state-of-the-art performance.

\vspace{1mm}
\subsubsection{Temporal modeling across different time spans} We measure $\mathcal{J}\&\mathcal{F}$, FLOPs per frame, and GPU memory usage of our VLP-RVOS with different inference clip lengths. As shown in Table~\ref{Tab:window_size}, our VLP-RVOS exhibits stable performance with varying inference clip lengths (the variance is 0.019), validating its robustness to clip length and strong temporal modeling ability. We also observe that GPU memory usage increases but the FLOPs decrease as the clip length increases. Users can adjust the inference clip length based on the available hardware memory without worrying about performance degradation in real-world applications.

\begin{table*}[t]
\begin{center}
\setlength{\tabcolsep}{8pt}
\caption{\textbf{Experimental results on A2D/JHMDB-Sentences.} All the models are trained on the A2D-Sentences training set alone.}
\renewcommand{\arraystretch}{1.0}
\resizebox{0.975\textwidth}{!}{
\begin{tabular}{l|c|ccc|ccc}
\toprule
\multirow{2}{*}{Algorithms} &\textbf{}
  \multirow{2}{*}{\tabincell{c}{Visual Backbone}} &
  \multicolumn{3}{c|}{A2D-Sentences} & 
  \multicolumn{3}{c}{JHMDB-Sentences}\\
  &&
  \multicolumn{1}{c}{mAP (\%)} &
  \multicolumn{1}{c}{IoU$_{\rm Overall}$ (\%)} &
  \multicolumn{1}{c|}{IoU$_{\rm Mean}$ (\%)} &
  \multicolumn{1}{c}{mAP (\%)} &
  \multicolumn{1}{c}{IoU$_{\rm Overall}$ (\%)} &
  \multicolumn{1}{c}{IoU$_{\rm Mean}$ (\%)} \\
\midrule
LBDT-4~\cite{LBDT} & ResNet-50 & 47.2 & 70.4 & 62.1 & 41.1 & 64.5 & 65.8\\
TempCD~\cite{TempCD} & ResNet-50 & \textcolor{white}{0.}-- & 76.6 & 68.6 & -- & 70.6 & 69.6 \\
LoSh-R~\cite{LoSh} & Video-Swin-T & 50.4 & 74.3 & 66.6 & 40.7 & 71.6 & 71.3  \\
SOC~\cite{SOC} & Video-Swin-T & 50.4 & 74.7 & 66.9 & 39.7 & 70.7 & 70.1\\
ReferFormer~\cite{ReferFormer} & Video-Swin-S & 53.9 & 77.7 & 69.8 & 42.4 & 72.8 & 71.5 \\
OnlineRefer~\cite{onlinerefer} & Video-Swin-B & \textcolor{white}{0.}-- & 79.6 & 70.5 & -- & 73.5 & 71.9\\
\rowcolor{gray!20}
\textbf{Ours} (CLIP) & ViT-B/16 & 53.3 & 76.7 & 69.5 & 44.2 & 73.6 & 71.9\\
\rowcolor{gray!20}
\textbf{Ours} (VLMo) & VLMo-B & 53.9 & 78.5 & 72.7 & 44.6 & 73.7 & 72.3\\
\rowcolor{gray!20}
\textbf{Ours} (CLIP) & ViT-L/14 & \underline{59.4} & \underline{84.0}   & \underline{75.3} & \underline{46.0} & \underline{77.9} & \underline{75.9}\\
\rowcolor{gray!20}
\textbf{Ours} (VLMo) & VLMo-L & \textbf{63.1} & \textbf{86.2} & \textbf{77.7} & \textbf{47.1} & \textbf{78.3} & \textbf{76.6} \\
\bottomrule
\end{tabular}}
\label{Tab:A2D}
\vspace{-2mm}
\end{center}
\end{table*}

\subsection{Comparison with state-of-the-art methods}
\noindent\textbf{Ref-Youtube-VOS \& Ref-DAVIS17.} Many RVOS methods use Video-Swin as the visual backbone, while ours use ViT as the visual encoder. For relatively fair evaluation, we measure the FLOPs per frame and speed of the RVOS models on RTX3090 and split those with \textbf{similar efficiency} to the same group for comparison. Specifically, we compare our models using ViT-B and VLMo-B with those using Video-Swin-T, and compare our models using ViT-L and VLMo-L with those using Video-Swin-B and Swin-L. Table~\ref{Tab:YTB} reports the results using different training protocols.

On Ref-Youtube-VOS, our models with VLMo achieve the best performance in all metrics in both groups, and our models with CLIP perform comparably with state-of-the-art RVOS methods in the two groups. These comparisons demonstrate the effectiveness of our VLP-RVOS framework. Besides, our models with VLMo and CLIP exhibit substantial advantages on Ref-DAVIS17 compared with other RVOS algorithms. The cross-dataset evaluation, \ie~training on Ref-Youtube-VOS and testing on Ref-DAVIS17, highlights the strong generalization ability of our VLP-RVOS. In terms of efficiency, our models achieve real-time or nearly real-time speeds.

\vspace{1mm}
\noindent\textbf{A2D-Sentences \& JHMDB-Sentences.} Table~\ref{Tab:A2D} presents the experimental results on A2D-Sentences and JHMDB-Sentences. All the models are trained on A2D-Sentences. Our models using ViT-B/14 CLIP and VLMo-B perform favorably against recently proposed methods using Video-Swin-T/S, such as Losh-R~\cite{LoSh}, SOC~\cite{SOC} and ReferFormer~\cite{ReferFormer}. Besides, our models using ViT-L/14 CLIP and VLMo-L outperform OnlineRefer~\cite{onlinerefer} using Video-Swin-B by large margins in IoU.

\begin{table}[t]
\begin{center}
\setlength{\tabcolsep}{10pt}
\caption{\textbf{Experimental results on the MeViS validation set.} All the models are trained on the MeViS training set alone.}
\renewcommand{\arraystretch}{1.0}
\resizebox{0.975\linewidth}{!}{
\begin{tabular}{l|ccc}
\toprule
Algorithms & \multicolumn{1}{c}{$\mathcal{J}\&\mathcal{F}$ (\%)} &
  \multicolumn{1}{c}{$\mathcal{J}$ (\%)} &
  \multicolumn{1}{c}{$\mathcal{F}$ (\%)} \\
\midrule
URVOS~\cite{URVOS} & 27.8 & 25.7 & 29.9\\
LBDT~\cite{LBDT} & 29.3 & 27.8 & 30.8\\
MTTR~\cite{MTTR} & 30.0 & 28.8 & 31.2\\
ReferFormer~\cite{ReferFormer} & 31.0 & 29.8 & 32.2\\
VLT+TC~\cite{vlt} & 35.5 & 33.6 & 37.3\\
LMPM~\cite{MeViS} & 37.2 & 34.2 & 40.2 \\
DsHmp~\cite{DsHmp} & \textbf{46.4} & \textbf{43.0} & \textbf{49.8}\\
\rowcolor{gray!20}
\textbf{Ours} (ViT-B/16 CLIP) & 44.6 & 41.3 & 48.0 \\
\rowcolor{gray!20}
\textbf{Ours} (VLMo-B) & \underline{45.4} & \underline{42.0} & \underline{48.8} \\
\bottomrule
\end{tabular}}
\label{Tab:Mevis}
\vspace{-2mm}
\end{center}
\end{table}

\vspace{1mm}
\noindent\textbf{MeViS.}
MeViS~\cite{MeViS} is a benchmark requiring RVOS models to understand the motion in video to locate and segment the target object. We conduct experiments on MeViS to evaluate the motion modeling ability of VLP-RVOS. Table~\ref{Tab:Mevis} reports the results. LMPM~\cite{MeViS} and DsHmp~\cite{DsHmp} are two RVOS algorithms elaborated to comprehend the motion of the target object with a motion perception mechanism, which achieve astonishing progress on MeViS. Although without explicit motion modeling at the object level, our models obtain comparable performance with DsHmp and better performance than LMPM, demonstrating its effectiveness in modeling the temporal context within the video clip.

\begin{figure*}[h!]
\centering
\scriptsize
\renewcommand\arraystretch{1.0}
\includegraphics[width=0.825\textwidth]{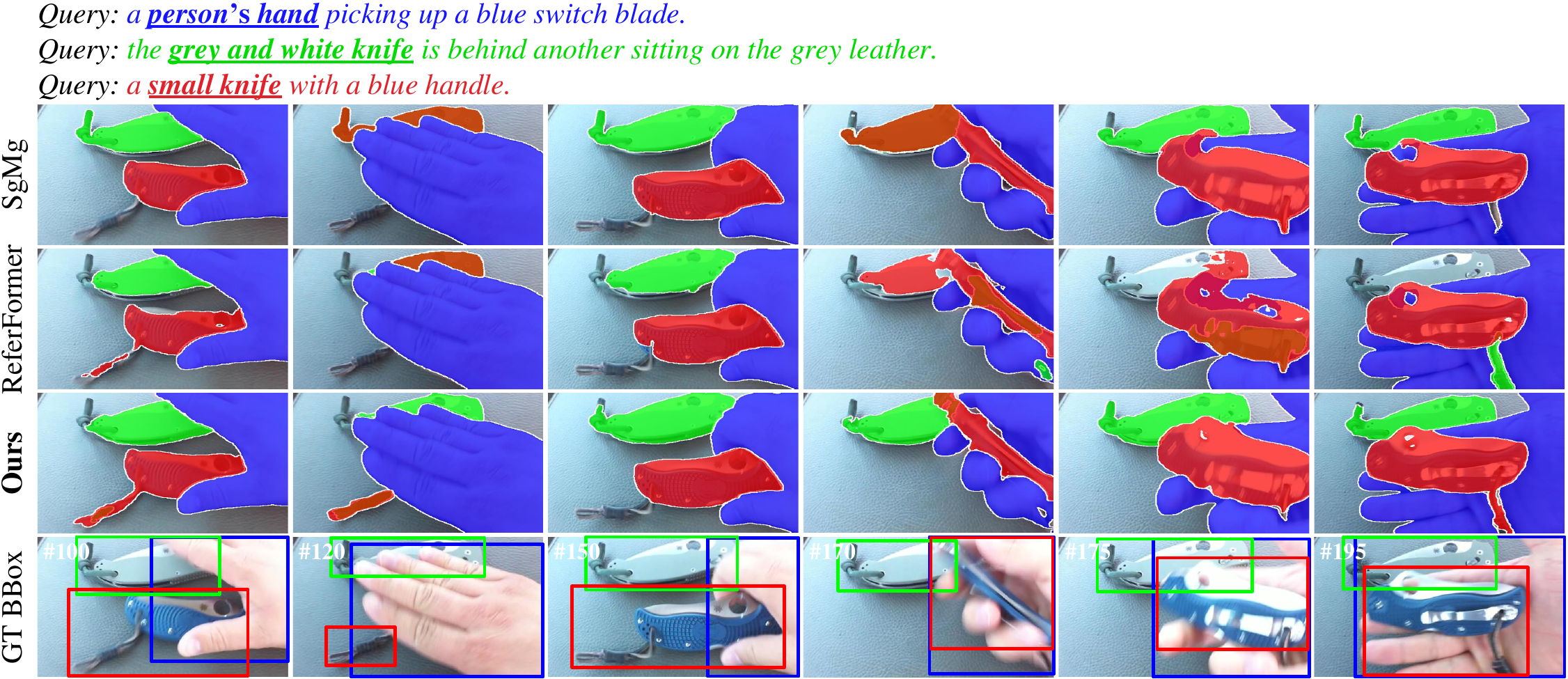}
\vspace{-3mm}
\caption{\textbf{Qualitative comparison between VLP-RVOS (VLMo-L), SgMg (Video-Swin-B), and ReferFormer (Video-Swin-B) on a video where a hand is picking up a knife.} All the methods can precisely segment the targets according to the descriptions at the beginning. Nevertheless, when the knives are occluded, SgMg and ReferFormer confuse the two similar knives, leading to erroneous predictions at the $120^{th}$ and $170^{th}$ frames. By contrast, our method is more robust to the occlusion and keeps segmenting the knives precisely.}
\label{Fig:Knife}
\end{figure*}

\begin{figure*}[h!]
\centering
\scriptsize
\renewcommand\arraystretch{1.0}
\vspace{-3mm}
\includegraphics[width=0.825\textwidth]{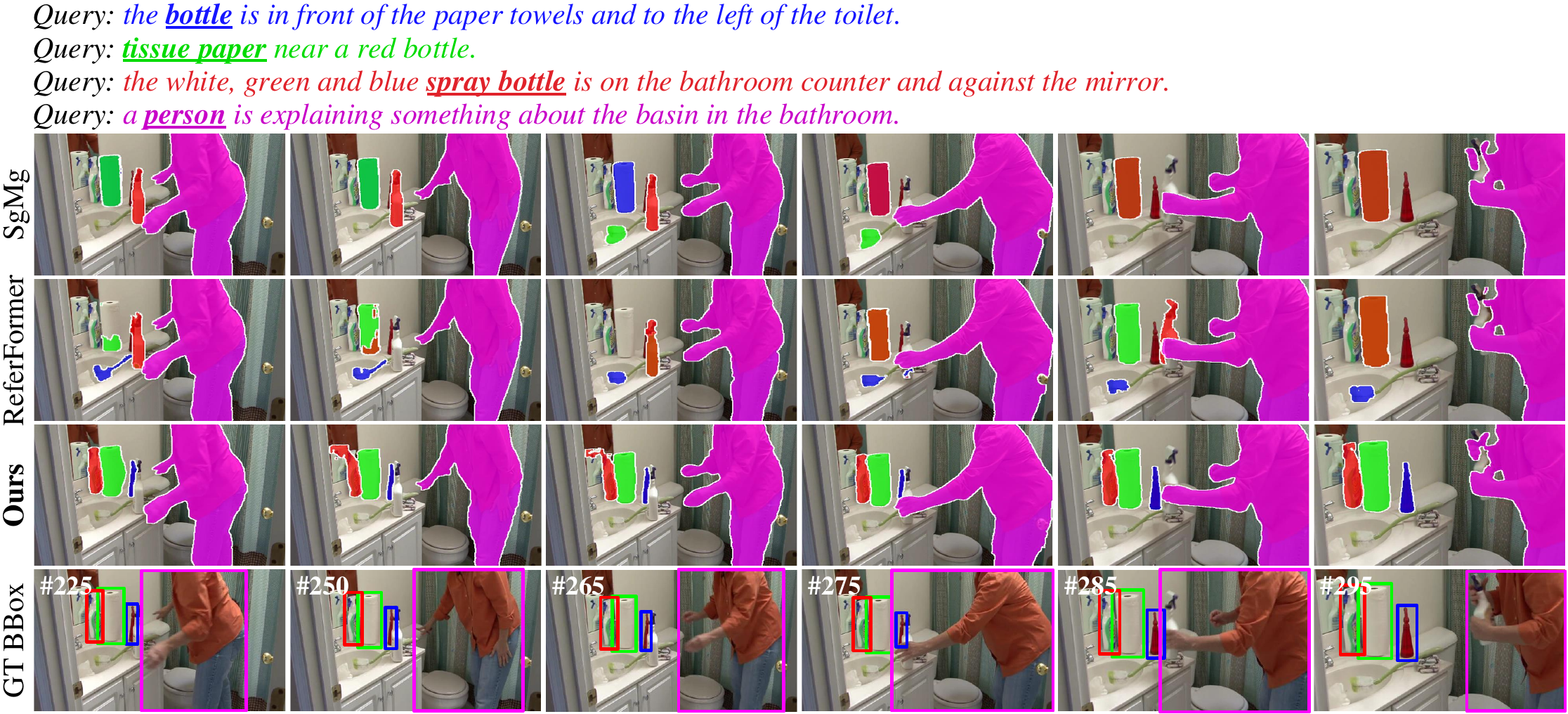}
\caption{\textbf{Qualitative comparison between VLP-RVOS (VLMo-L), SgMg (Video-Swin-B), and ReferFormer (Video-Swin-B) on a video where some cleaning supplies are placed on the sink.} Our method accurately comprehends the spatial positions and appearances described in the queries, successfully locating these cleaning supplies. In contrast, SgMg and ReferFormer encounter difficulties in understanding the descriptions within this cluttered scene.}
\label{Fig:Bottles}
\end{figure*}

\subsection{Qualitative results}
we present qualitative results on several challenging videos to obtain more insights into the pros and cons of VLP-RVOS.

\vspace{1mm}
\noindent\textbf{Comparisons with state-of-the-art methods.}
We first qualitatively compare our VLP-RVOS (VLMo-L) with two state-of-the-art algorithms ReferFormer (Video-Swin-B) and SgMg (Video-Swin-B) on two videos. Figure~\ref{Fig:Knife} illustrates the segmentation results in a video where the referred objects undergo occlusions. Our VLP-RVOS exhibits superior robustness compared to ReferFormer and SgMg. The favorable performance manifests that the temporal-aware prompt-tuning method and the spatial-temporal reasoning module equip the model with strong temporal modeling ability. 

Figure~\ref{Fig:Bottles} shows the segmentation results in a video with cluttered scenes. Our VLP-RVOS accurately comprehends detailed descriptions and precisely segments the target objects, whereas ReferFormer and SgMg face difficulties in understanding these complex scenarios. The favorable performance shows that transferring the knowledge of VLP models boosts the vision-language understanding ability of our model.

\begin{figure*}[h!]
\centering
\includegraphics[width=0.85\textwidth]{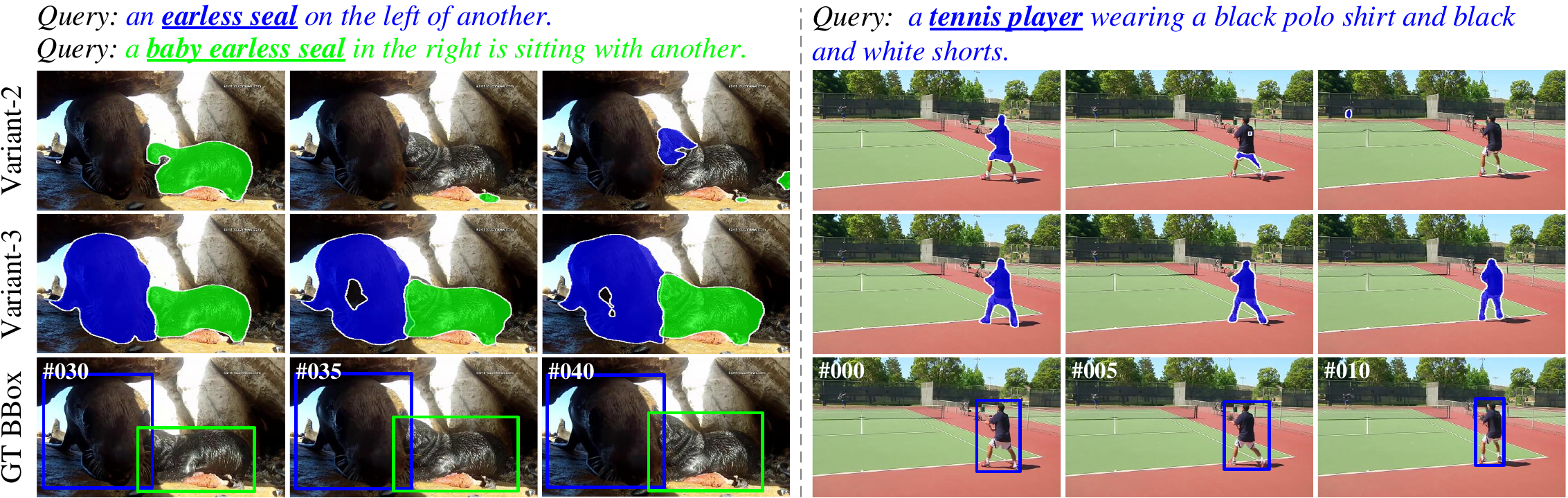}
\vspace{-3mm}
    \caption{\textbf{Qualitative comparisons between Variant-2 (w/o temporal prompts) and Variant-3 (w/ temporal prompts) on two challenging videos.} For each video, we visualize the prediction results on the consecutive frames of the same clip to analyze the effect of the temporal prompt. Without considering the temporal clues, Variant-2 performs segmentation on each frame independently. Consequently, it predicts inconsistent masks across consecutive frames within a video clip. By contrast, Variant-3 with temporal prompts generates more stable and consistent predictions on consecutive frames.}
\label{Fig:temporal_prompt}
\vspace{-2mm}
\end{figure*}

\begin{figure*}[h!]
\centering
\includegraphics[width=0.875\textwidth]{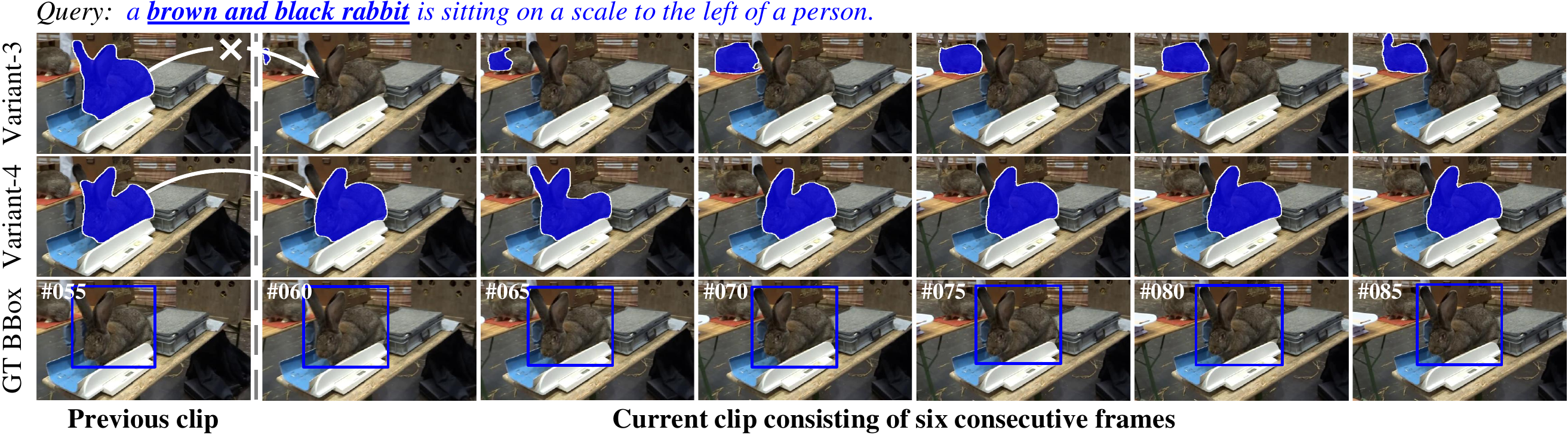}
\vspace{-3mm}
    \caption{\textbf{Qualitative comparisons between Variant-3 (w/o historical prompts) and Variant-4 (w/ historical prompts) on a challenging video.} The $55^{th}$ frame is from the previous clip, and both Variant-3 and Variant-4 successfully locate the target rabbit in this frame. With the historical prior of the target rabbit, Variant-4 keeps tracking it in the current clip (from the $60^{th}$ to the $85^{th}$ frame). By contrast, the segmentation masks of Variant-3 drift to the distractor in the current clip, which is a similar rabbit gradually appearing in the view.}
\label{Fig:historical_prompt}
\vspace{-2mm}
\end{figure*}

\begin{figure}[h!]
\centering
\includegraphics[width=1.0\linewidth]{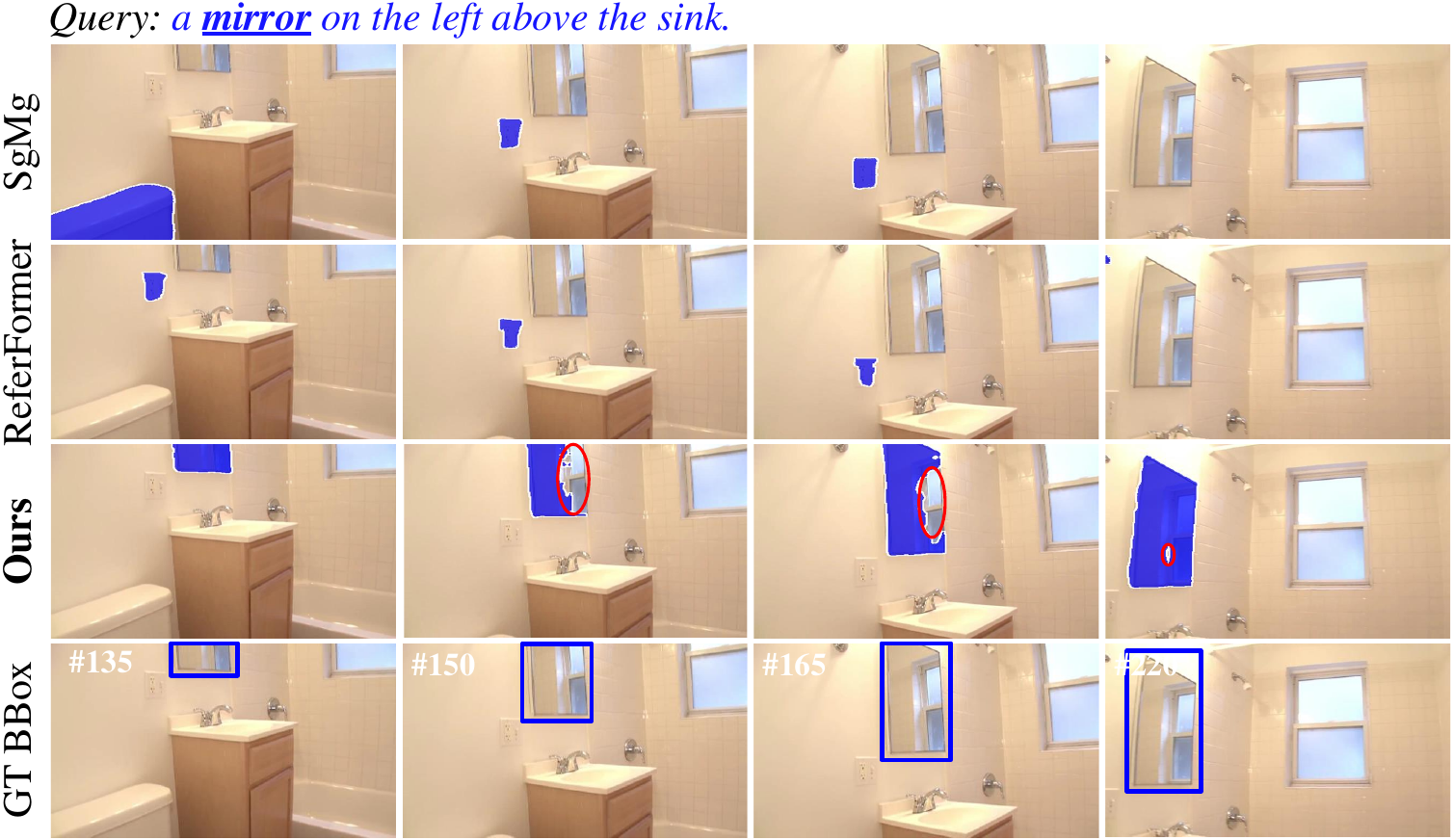}
\vspace{-6mm}
    \caption{\textbf{Prediction results for segmenting a mirror.} The target object is a mirror located on the left above the sink. Although being able to understand the concept of the mirror and locate it, VLP-RVOS has difficulty distinguishing between the reflection inside the mirror and the real object outside, as highlighted by the red circles.}
    \vspace{-2mm}
\label{Fig:mirror_case}
\end{figure}

\vspace{1mm}
\noindent\textbf{Comparison between w/o and w/ temporal prompts.}
We qualitatively compare Variant-2 (w/o temporal prompts) and Variant-3 (w/ temporal prompts) to obtain more insights into the effect of the temporal prompt. Figure~\ref{Fig:temporal_prompt} illustrates their segmentation results over consecutive frames of the same clip on several videos. Note that the videos in Ref-Youtube-VOS are annotated every 5 frames and VLP-RVOS performs segmentation on the annotated frames. We can observe that Variant-3 generates much more stable and consistent segmentation masks across consecutive frames than Variant-2. These comparisons demonstrate the effectiveness of the temporal prompts for temporal modeling.

\vspace{1mm}
\noindent\textbf{Comparison between w/o and w/ historical prompts.}
We also qualitatively compare Variant-3 (w/o historical prompts) and Variant-4 (w/ historical prompts) to analyze the effect of the historical prompt. Figure~\ref{Fig:historical_prompt} illustrates their segmentation results over two consecutive clips on a video. Both Variant-3 and Variant-4 can locate the target rabbit in the previous clip (as shown in the $55^{th}$ frame). Nevertheless, Variant-3 loses the referred rabbit and drifts to a distractor in the following clip (from the $60^{th}$ to the $85^{th}$ frame). By contrast, Variant-4 with the target prior from the previous clip continues tracking this rabbit in the following clip.

\vspace{1mm}
\noindent\textbf{Failure cases.}
Although VLP-RVOS has shown promising vision-language understanding abilities in the above experiments, we observe a challenge in distinguishing between reflections inside mirrors and real objects outside. As shown in Figure~\ref{Fig:mirror_case}, VLP-RVOS can recognize the presence of the target mirror, but it incorrectly identifies the reflection inside the mirror as the real object. A potential and straightforward solution is to further enhance the contextual modeling ability and meanwhile incorporate the mirror data~\cite{Mirrordata} for training.

\section{Conclusion}
We have presented a VLP-RVOS framework to transfer VLP models to RVOS. It enables learning relation modeling for RVOS from aligned VL space instead of from scratch. Specifically, we propose a temporal-aware prompt-tuning method, which not only adapts pre-trained representations for pixel-level prediction but also empowers the vision encoder to model temporal clues. We further design a cube-frame attention mechanism for efficient and effective spatial-temporal reasoning. Besides, we propose a multi-stage VL relation modeling scheme for comprehensive VL understanding. Extensive experiments on four benchmarks demonstrate the effectiveness and generalization of VLP-RVOS.

\bibliographystyle{IEEEtran}
\bibliography{main}

\end{document}